\documentclass[runningheads]{llncs}

\usepackage[mobile]{eccv}

\usepackage{eccvabbrv}
\usepackage{graphicx}
\usepackage{booktabs}
\usepackage[accsupp]{axessibility}
\usepackage{hyperref}
\usepackage{orcidlink}

\begin{document}

\title{Decoupled Illumination Priors for Spatially Controllable Multi-View Indoor Scene Relighting} 

\titlerunning{Lume-Palette}

\author{Chenjian Gao\inst{1}\orcidlink{0009-0003-6327-4893} \and
Linning Xu\inst{1}$^\dag$\orcidlink{0000-0003-1026-2410} \and
Tianfan Xue\inst{1,2,3}$^\dag$\orcidlink{0000-0001-5031-6618}}

\authorrunning{C.~Gao et al.}

\institute{Multimedia Laboratory, The Chinese University of Hong Kong \and
Shanghai AI Laboratory \and CPII under InnoHK \\
\email{\{gc025, tfxue\}@ie.cuhk.edu.hk} \quad \email{linningxu@link.cuhk.edu.hk} \\
\mbox{$^\dag$ Corresponding Authors}} 

\maketitle
\vspace{-3mm}
\begin{center}
{\href{https://cjeen.github.io/lumepalette}{https://cjeen.github.io/lumepalette}}
\end{center}

\begin{abstract}
Indoor scene relighting demands photorealism, precise spatial control, and strict multi-view consistency. While diffusion-based image editing models enable semantic lighting manipulation via text prompts, enforcing exact 3D light placement often disrupts their generative priors.
We propose Lume-Palette, a progressive framework that leverages semantic lighting priors for spatially controllable multi-view indoor relighting. The approach decouples relighting into two stages: (1) illumination distillation, which extracts canonical illumination palettes from a pretrained diffusion model to preserve realistic material–light interactions, and (2) illumination casting, which explicitly maps target spatial lighting conditions defined from coarse 3D geometry. To efficiently handle dense multi-view and multi-modal inputs, we introduce an asymmetric multi-view conditioning strategy that selectively injects essential spatial context.
Experiments on diverse synthetic scenes and real-world scenes demonstrate that Lume-Palette produces photorealistic, spatially controllable, and multi-view consistent relighting results.
\end{abstract}

\section{Introduction}
\begin{figure}[t]
\centering
\includegraphics[width=\linewidth]{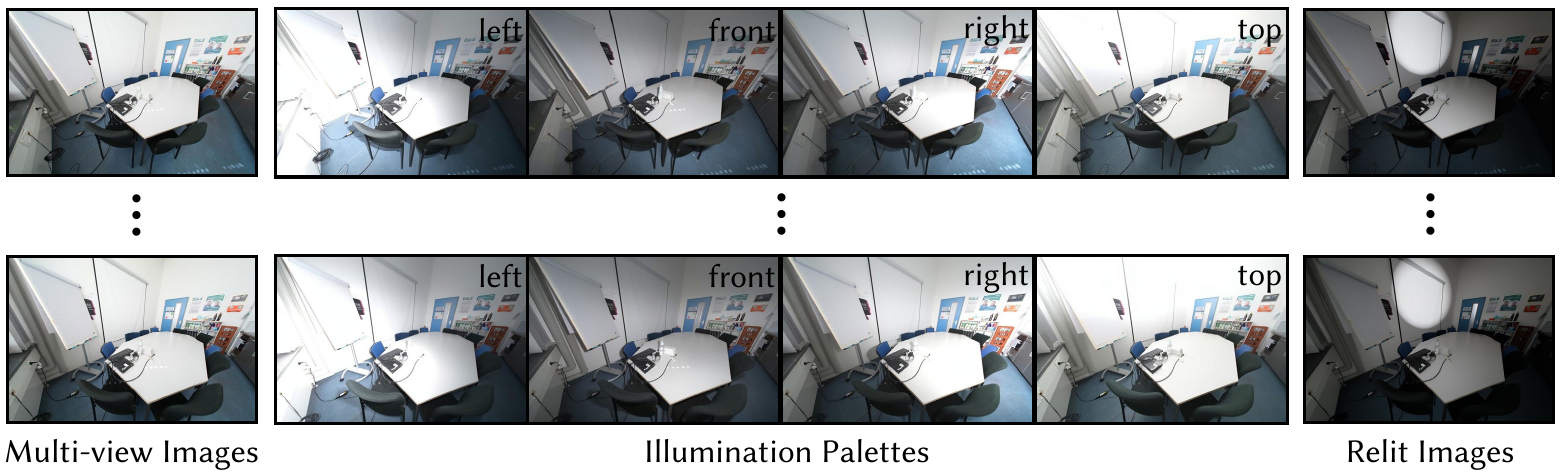}
\caption{Multi-view indoor scene relighting with Lume-Palette. Our progressive framework first extracts canonical "illumination palettes" from multi-view source images to capture material-light interactions. These palettes subsequently guide the synthesis of photorealistic, multi-view consistent relit images under new target illuminations.}
\label{fig:teaser}
\end{figure}
Indoor scene relighting stands as a pivotal challenge in computer vision and computer graphics, with applications ranging from immersive augmented reality (AR) and 3D virtual photography to interior design~\cite{lin2025iris, xing2025luminet, choi2025scribblelight, ji2025digital}. In these interactive environments, users explore the space from varying viewpoints, necessitating illumination that is not only realistic but also consistent across different viewpoints. Unlike object-centric relighting~\cite{litman2025lightswitch, zeng2024dilightnet, jin2024neural}, scene-level relighting presents a unique set of complexities, as the geometry is no longer isolated and observed from limited angles. It requires maintaining coherent global illumination and shadow propagation amidst intricate occlusion relationships and diverse material properties. Due to these complexities, while traditional inverse rendering methods~\cite{bell2001view, barron2014shape} strive for physical interpretability, they often struggle to accurately invert material interactions, yielding results that lack photorealism.

Recently, diffusion-based image editing models~\cite{batifol2025flux, wu2025qwen, flux-2-2025} have excelled at manipulating lighting effects via text prompts, yielding photorealistic results. However, text-driven control lacks explicit spatial control over light sources and struggles to preserve consistency when the viewpoint shifts. To achieve spatial lighting control and multi-view consistency, recent approaches~\cite{liang2026luxremix, litman2025lightswitch} attempt to fine-tune diffusion models to accept spatial condition maps. Forcing models to accommodate these invasive spatial constraints from scratch by relying on synthetic data can disturb their generative priors. Consequently, while these approaches achieve spatial alignment, they tend to produce synthetic artifacts, failing to balance precise illumination controllability with natural realism.

Our goal is to achieve explicit spatial control for multi-view consistent relighting without losing the diffusion model's native lighting prior. Drawing inspiration from photometric stereo~\cite{goldman2009shape, debevec2000acquiring}, which reveals material properties by capturing a scene under a set of canonical lighting directions, we propose \textbf{Lume-Palette}, a progressive framework that decouples the relighting process into illumination distillation and illumination casting.
Since pre-trained diffusion models are not natively trained to condition on explicit spatial lighting maps, forcing them to condition on this foreign modality requires learning a complex geometry-to-appearance mapping from scratch. This heavy adaptation inevitably disrupts the model’s well-tuned generative priors, degrading photorealism. To bypass this, we first distill canonical "illumination palettes" via text prompts—a modality the model is inherently aligned with—preserving authentic material responses. These high-quality references then allow the casting stage to focus solely on spatial distribution guided by the geometric conditions. Directionally relit images have also been used by Poirier-Ginter~\etal~\cite{poirier2024diffusion} as priors for per-scene relightable radiance-field optimization. In Lume-Palette, these priors are instead distilled into illumination palettes and cast through a feed-forward diffusion relighting engine, enabling spatially controllable 3D lighting across multiple views.

In the illumination distillation stage, we distill the generative prior of a pretrained diffusion model into a photorealistic \textbf{illumination palette}—a set of reference images capturing the scene under canonical lighting directions (e.g., left, right, front, and top), as illustrated in Fig.~\ref{fig:teaser}.
By fine-tuning the model on real-world photographs to align with fixed lighting descriptive prompts, we induce the synthesis of these canonical lighting patterns. This semantic-driven distillation avoids invasive conditioning, preserving the rich generative prior. By revealing how the scene's surfaces respond to canonical illumination, they serve as a rich generative prior to guide the synthesis of highly photorealistic results. 

In the subsequent illumination casting stage, we leverage the illumination palette to synthesize target illumination. Specifically, we allow users to freely place virtual light sources within a reconstructed coarse 3D mesh, and then render the scene's response to represent the target lighting conditions.
Unlike explicit source parameterization~\cite{magar2025lightlab}, our adopted receiver-centric lighting condition naturally accommodates off-screen lights while maintaining cross-view 3D consistency.
To synthesize the final relit scene, the network must jointly process the multi-view source images alongside their corresponding lighting conditions and the previously distilled illumination palettes. However, simply concatenating all these dense, multi-modal conditions across all viewpoints is computationally intractable. 
To overcome this, we design an asymmetric multi-view conditioning scheme that avoids feeding dense conditions to every view. 
For each prediction, only the active view receives the full source image, lighting condition, and illumination palette, while the other views are represented by lightweight noisy latents as spatial anchors, enabling scalable joint processing for multi-view relighting.

Our approach achieves photorealistic, spatially controllable, and multi-view consistent relighting results by leveraging the generative priors of powerful image generative models via illumination distillation and illumination casting. Experiments on synthetic and real-world scenes show that Lume-Palette produces visually plausible relighting results that follow user-specified spatial lighting conditions while maintaining cross-view coherence.
\section{Related Works}
\label{sec:related}

\subsection{Inverse Rendering and Physics-Based Relighting}
Inverse rendering traditionally decomposes images into intrinsic components via optimization with hand-crafted priors~\cite{barron2014shape, lombardi2019neural, bell2001view, grosse2009ground}. Deep learning advanced this by directly predicting SVBRDFs and lighting~\cite{li2020inverse, sengupta2019neural, deschaintre2018single, roberts2021hypersim}, often utilizing vision transformers for dense estimation~\cite{zhu2022irisformer, wang2021single}. To ensure physical plausibility, differentiable rendering~\cite{azinovic2019inverse, nimier2019mitsuba, li2022physically, careaga2025physically} and neural implicit representations~\cite{boss2021nerd, zhang2021physg, srinivasan2021nerv, bi2020deep} are widely adopted to jointly optimize scene parameters and complex reflectance fields. However, explicit inversion remains computationally expensive and highly sensitive to geometric inaccuracies, often producing unnatural artifacts under complex global illumination. Instead of relying on explicit decomposition, our approach uses illumination palettes as intermediate appearance references that provide material-dependent lighting cues for spatially controllable relighting.

\subsection{Generative and Diffusion-Based Relighting}
While GANs laid the foundation for neural relighting~\cite{wang2019underexposed, sun2019single, zhou2019deep, pandey2021total}, diffusion models dominate image editing~\cite{rombach2022high, saharia2022palette, zhang2023controlnet}. Text prompts manipulate lighting semantics~\cite{brooks2023instructpix2pix, hertz2022prompt, mou2024t2i}, but lack spatial precision; recent methods therefore add spatial maps~\cite{zeng2024rgb, kim2024switchlight}, light-transport constraints~\cite{zhang2024iclight, zeng2024dilightnet}, or user guidance~\cite{choi2025scribblelight, magar2025lightlab, liang2026luxremix}. Other systems explore lighting-aware editing or control from complementary perspectives, including intrinsic-space editing~\cite{lyu2025intrinsicedit}, unified geometry--illumination diffusion for controllable video~\cite{lin2026illumicraft}, and practical light-control interfaces~\cite{erel2025practilight}. These approaches highlight explicit control, but adapting diffusion models to unfamiliar spatial modalities can disturb the pre-trained generative prior.

To preserve generative relighting priors, recent work distills lighting knowledge by isolating illumination semantics~\cite{kocsis2024lightit}, extracting environment maps via LoRA~\cite{phongthawee2024diffusionlight}, or distilling ray tracing into video models~\cite{bharadwaj2025genlit}. Directionally relit images have also been used as intermediate priors: Poirier-Ginter~\etal~\cite{poirier2024diffusion} fine-tune a 2D diffusion model on MIT Multi-Illumination with global light-direction conditioning, and use the generated relit images for per-scene radiance-field optimization; ROGR~\cite{tang2025rogr} uses generative relighting for relightable 3D objects. Lume-Palette follows this line of using generated relighting results as illumination priors, but casts canonical relit images through receiver-centric spatial lighting conditions for feed-forward multi-view indoor relighting with user-placed 3D lights, without reconstructing a relightable radiance field per scene.

Rendered cues are another established control signal: DiffusionRenderer~\cite{liang2025diffusion} uses G-buffers, LightIt~\cite{kocsis2024lightit} uses shading maps, and DiFaReli~\cite{ponglertnapakorn2023difareli}, Careaga and Aksoy~\cite{careaga2025physically}, and Hybrelighter~\cite{zhao2025hybrelighter} use face, mesh, or reconstructed-scene renderings. We also render a receiver-centric lighting cue, but use it as a spatial layout rather than the sole appearance driver: it provides 3D-consistent guidance for the target illumination distribution, while the palette supplies view-specific material-response cues for the casting network.

\subsection{Multi-View Consistent Relighting}
Multi-view relighting must preserve consistency despite occlusions and viewpoint changes. Warping~\cite{philip2019multiview, philip2021free} is brittle, and neural fields~\cite{zhang2021nerfactor, boss2021nerd, rudnev2022nerf, gao2024relightable3dgs} require per-scene optimization. 
Generative methods use video attention~\cite{he2025unirelight, xing2025luminet}, joint denoising~\cite{shi2024mvdream, liu2024syncdreamer, gao2024genesistex}, propagation, including GS-Light~\cite{ye2025training}, or harmonization~\cite{liang2026luxremix, trevithick2025simvs}.
LightSwitch~\cite{litman2025lightswitch} avoids all-pair attention via mini-batch view shuffling for object relighting. While it samples different view subsets with similar per-view conditioning, our setting involves much denser conditions for each scene view. We therefore use asymmetric conditioning: only the active view receives these dense conditions, while inactive views are kept as lightweight latent anchors to provide cross-view cues through the shared latent state.

\section{Method}
\begin{figure}[t]
    \centering
    \includegraphics[width=1.0\linewidth]{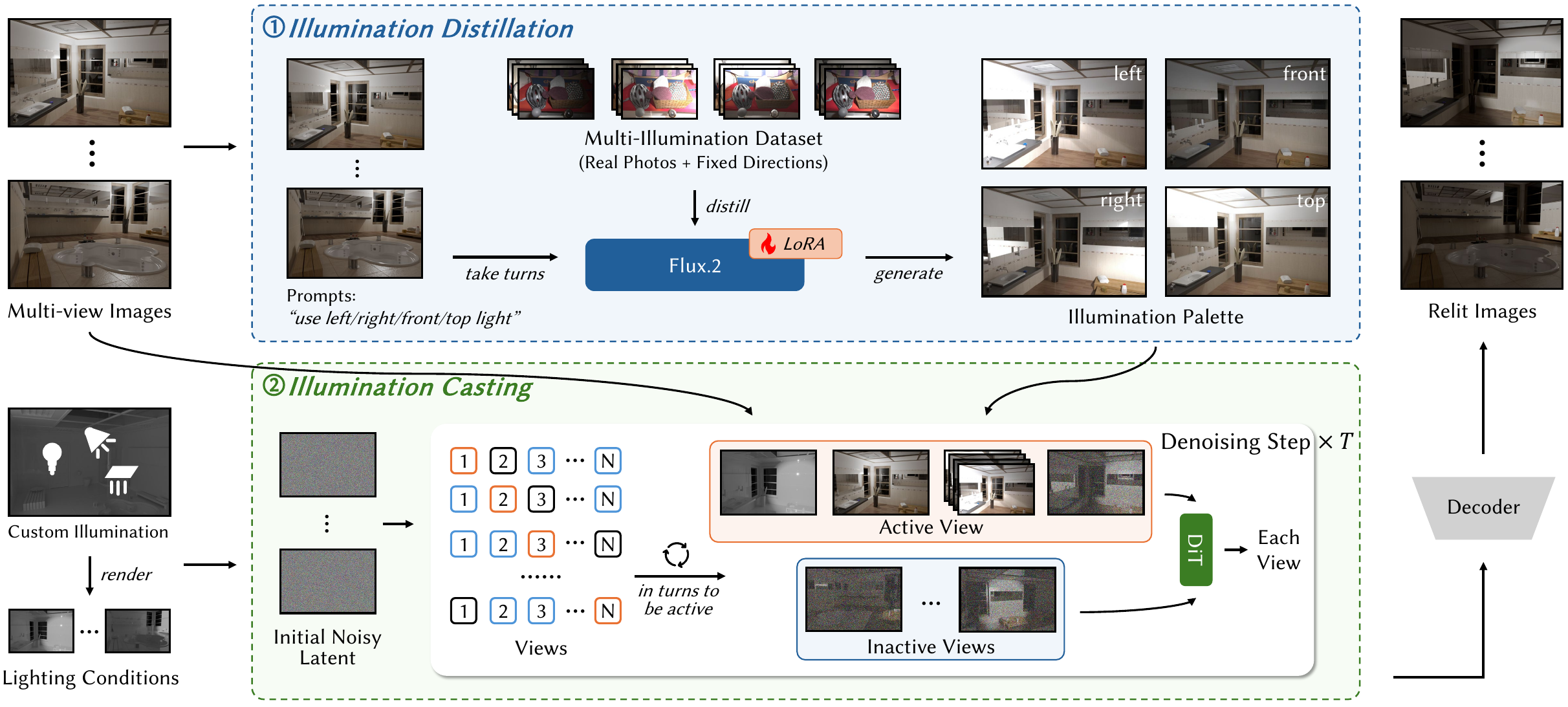}
    \caption{Overview of Lume-Palette. Our framework decouples indoor relighting into two sequential stages. \textit{Illumination Distillation} fine-tunes a diffusion model to extract canonical illumination palettes, capturing intrinsic material responses. \textit{Illumination Casting} synthesizes the target illumination by combining these palettes with source images and user-defined spatial lighting conditions. To ensure scalable multi-view consistency, this stage employs an asymmetric multi-view conditioning scheme.}
    \label{fig:pipeline}
\end{figure}
\subsection{Overview}
\label{sec:overview}

We present \textbf{Lume-Palette}, a framework designed to synthesize photorealistic and spatially consistent indoor relighting results.
Formally, given a set of $N$ multi-view source images $\mathcal{I} = \{I_1, \dots, I_N\}$ captured in the same scene under consistent lighting conditions, and a globally consistent user-defined spatial lighting condition represented as view-specific spatial lighting maps $\mathcal{L} = \{L_1, \dots, L_N\}$, our framework predicts multi-view consistent relit images $\mathcal{R} = \{R_1, \dots, R_N\}$. 
Here, each $R_i$ is the relit version of the corresponding source image $I_i$, faithfully reflecting the target illumination. 
A naive approach to solve this problem is a single-stage pipeline that directly conditions an image editing model on explicit spatial lighting conditions. 
However, this end-to-end paradigm faces two critical challenges. First, in the real world, it is very hard to capture multi-view consistent and spatially varying relight pairs with precise spatial lighting annotations. Consequently, models must rely heavily on synthetic datasets, which inevitably compromises the photorealistic generative priors of the pretrained model. 
Second, it is hard to train a single network to simultaneously erase complex lighting in the original input and render complex target illumination, as it requires the network to learn intrinsic image decomposition, lighting estimation, and geometry estimation jointly without intermediate supervision.

To address both challenges, our approach decouples the relighting process into two sequential stages: \textbf{illumination distillation} and \textbf{illumination casting} (Fig.~\ref{fig:pipeline}).
The distillation stage learns a generative prior for canonical lighting directions from real-world photographs and their corresponding directional annotations, which are simple textual descriptions indicating the light source positions (e.g., `left'', `right'').
By distilling this prior into specialized Low-Rank Adaptation (LoRA) modules, the model can generate images with canonical directional illumination for a given source image, serving as an intermediate illumination palette for the following lighting stage.
The casting stage then synthesizes relighting results under specific lighting conditions. This is achieved by combining the original multi-view images and these reference palettes, guided by spatial lighting conditions.
The lighting condition includes multi-view shading images generated by a rendering engine; it encourages multi-view consistent relighting through shared 3D-derived conditions.
Consequently, our approach preserves the photorealistic prior by decoupling the native generative lighting prior from the enforcement of specific lighting control.
Furthermore, by normalizing the complex source illumination with basic canonical lighting, the intermediate palettes alleviate the burden of erasing the original lighting during the casting stage.

\subsection{Illumination Distillation}
\label{sec:semantic_distillation}

As previous diffusion-based relighting methods show, diffusion models have an implicit prior regarding light-material interactions. However, since this prior is only implicit, it does not support flexible control and multi-view consistent relighting. Therefore, we first distill this prior into an explicit representation of light-material interactions, named the \textbf{illumination palette}. This palette is defined as the four lighting results of an input image lit from 4 canonical lighting directions $d \in \{\text{`left'}, \text{`right'}, \text{`front'}, \text{`top'}\}$. To achieve this, we train specialized Low-Rank Adaptation (LoRA)~\cite{hu2022lora} modules of a pretrained diffusion model to generate this palette. During inference, these trained LoRA modules are applied independently to each individual image $I_i$ within the multi-view source set $\mathcal{I} = \{I_1, \dots, I_N\}$. Guided by text prompts, this process generates a view-specific illumination palette $\mathcal{B}_i = \{ B_i^{left}, B_i^{right}, B_i^{front}, B_i^{top} \}$ for every viewpoint $i$, consisting of reference images capturing the scene's authentic shading response to the distinct canonical directions $d$.

To supervise this distillation, we leverage the MIT Multi-Illumination~\cite{murmann19} dataset, which features 1,000 indoor scenes.
For each scene, the camera remains strictly fixed while capturing photographs under 25 distinct lighting directions, ensuring perfectly aligned multiple illumination image groups.
To construct our canonical illumination palette, we train four independent LoRA modules corresponding to target directions $d \in \{\text{`left'}, \text{`right'}, \text{`front'}, \text{`top'}\}$. These four LoRA modules are shared across all scenes.
During a training iteration of each LoRA (like $d=\text{`left'}$), we select that lighting direction as the ground-truth target $I_d$, and randomly sample one another lighting out of the 24 available lighting from the same scene in the dataset as the input image $I_{src}$.
By forcing the model to map these diversely lit inputs to the same fixed target, we ensure the network learns to override arbitrary initial illumination with the desired canonical lighting.
Formally, since our base model~\cite{flux-2-2025} operates on a flow matching framework~\cite{lipman2022flow}, we optimize the LoRA module $\theta$ by minimizing the velocity matching objective. Let $\mathbf{z}_0 = \mathcal{E}(I_d)$ be the latent representation of the target image encoded by a pretrained VAE, and $\mathbf{z}_1 \sim \mathcal{N}(\mathbf{0}, \mathbf{I})$ be sampled Gaussian noise. We construct the time-dependent noisy state as $\mathbf{z}_t = (1-t)\mathbf{z}_0 + t\mathbf{z}_1$. The optimization objective is defined as:
$$
    \min_{\theta} \left\| v_\theta(\mathbf{z}_t, t, \mathcal{E}(I_{src}), \text{``use } d \text{ light''}) - (\mathbf{z}_1 - \mathbf{z}_0) \right\|^2
$$
where $v_\theta$ represents the velocity prediction network equipped with the LoRA module $\theta$, and $t \in [0, 1]$ is the time step.

At inference, given a set of multi-view source images $\mathcal{I} = \{I_1, \dots, I_N\}$, we apply the four fine-tuned LoRA modules to each viewpoint independently to predict the scene's appearance under the canonical directions. The resulting view-specific collection of illumination palettes, denoted as $\mathcal{B} = \{ B_i^{left}, B_i^{right}, B_i^{front}, B_i^{top} \}_{i=1}^N$, serves as lighting references that capture how scene materials respond to canonical illumination, such as specularities and shading falloff, ready for the subsequent illumination casting stage.

\subsection{Illumination Casting}
\begin{figure}[t]
\centering
\includegraphics[width=\linewidth]{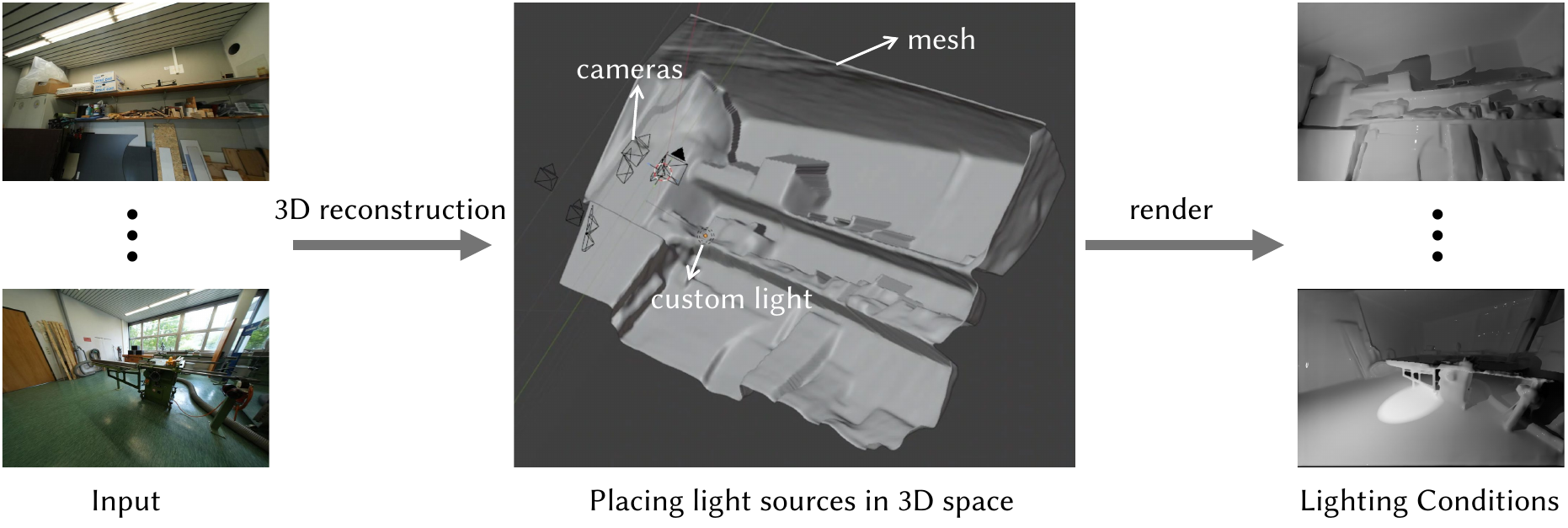}
\caption{Generation of spatial lighting conditions. First, a coarse 3D mesh is reconstructed from multi-view input images\cite{depthanything3}. Users can then explicitly place custom light sources within this 3D space. Finally, the scene is rendered to produce view-specific lighting conditions for the illumination casting stage.}
\label{fig:gui}
\end{figure}
\label{sec:casting}
While the illumination palettes provide robust material cues, they lack precise light control due to the ambiguity of text prompts.
The goal of the illumination casting stage is to combine these palettes with explicit spatial lighting conditions to synthesize consistent multi-view relighting under user-defined light sources.
Formally, the input to this stage includes the multi-view source images $\mathcal{I} = \{I_1, \dots, I_N\}$, the distilled illumination palettes $\mathcal{B} = \{ B_i^{left}, B_i^{right}, B_i^{front}, B_i^{top} \}_{i=1}^N$ estimated in the previous illumination distillation, and the target spatial lighting conditions $\mathcal{L} = \{L_1, \dots, L_N\}$.
By conditioning the model on these combined inputs, this stage outputs the final synthesized relit images $\mathcal{R} = \{R_1, \dots, R_N\}$.
As detailed below, $\mathcal{L}$ expresses the specific distribution of illumination by allowing users to freely place various light sources within the 3D space, providing more precise control than abstract text prompts.

\subsubsection{Spatial Lighting Condition}

To explicitly define the target spatial illumination, we construct the lighting condition images in the rendering engine using the estimated scene 3D geometry.
First, given the multi-view source images $\mathcal{I} = \{I_1, \dots, I_N\}$, we utilize Depth Anything 3~\cite{depthanything3} to extract their corresponding depth maps and reconstruct a coarse 3D mesh of the scene.
Second, to support flexible control of illumination, we allow users to arbitrarily place and configure various light sources (such as point, spot, or area lights) within this reconstructed 3D space.
Details of the user interface are provided in the supplementary material.
Given the estimated geometry and customized lighting, we assign a uniform, untextured white material to the 3D mesh and render it from the $N$ target viewpoints under these user-defined lights (this is sometimes called a ``white 3D model''), yielding the view-specific spatial lighting conditions $\mathcal{L} = \{L_1, \dots, L_N\}$.

The main idea of this design is to use multi-view 2D rendering of an untextured surface, rather than the explicit 3D location or types of lighting sources, as the lighting condition. This offers two advantages. First, it supports arbitrary choices of lighting, including off-screen or complex lighting setups without requiring visible light sources. Second, because these conditions $\mathcal{L}$ are derived from a shared 3D representation, this approach provides 3D-consistent lighting conditions across views.

\subsubsection{Asymmetric Multi-View Conditioning}

Achieving multi-view consistency implies modeling the conditional joint probability distribution $p(\mathcal{R} \mid \mathcal{I}, \mathcal{L}, \mathcal{B})$.
The most direct approach would be to simultaneously condition the generation on the complete collection of inputs across all $N$ viewpoints within a single network pass, comprising all the condition images per view: the source image $I_i$, the spatial lighting condition $L_i$, and the four directional references $\{ B_i^{left}, B_i^{right}, B_i^{front}, B_i^{top} \}$.
However, concatenating such an extensive set of condition images leads to an unmanageable explosion in input dimensionality, making direct joint modeling computationally intractable.

To address this bottleneck while preserving spatial awareness, we formulate our approach as an asymmetric multi-view conditioning scheme. 
Since processing the full set of $N$ views is intractable, we sample a group of $V$ views ($V < N$, e.g., $N=8, V=4$) to serve as the local context window. 
Within this focused context window, we select one view index $m$ as an active view and other views as inactive views. 
The model is trained to predict the velocity field exclusively for this active view. 
The model uses the complete set of that active view as dense guiding signals: the source image of that active view as $I_m \in \mathcal{I}$, the target spatial lighting condition $L_m \in \mathcal{L}$, and the illumination palettes $\{B_m^{left}, B_m^{right}, B_m^{front}, B_m^{top}\}$ distilled in the first stage. 
Since the model does not predict the velocity field for these inactive views, feeding them dense visual conditions is unnecessary. 
Instead, to maintain cross-view spatial awareness between the active and inactive views, we solely incorporate their concurrent noisy latents $\{\mathbf{z}_t^k\}_{k \neq m}$ to serve as lightweight spatial anchors. 
These inactive views are necessary here because they provide the global 3D context needed to ensure 3D consistency during the denoising process.
Unlike the dense conditions that require encoding multiple high-dimensional reference images, these noisy latents are merely the intermediate noisy states of the relit images, making them computationally trivial. 
By concatenating these asymmetric signals, the final input stream encompasses the active view's dense conditions and the noisy latents of inactive views. 
By omitting the dense condition encodings for the $V-1$ inactive views, this asymmetric split reduces the condition sequences from $6V$ to $V+5$. For example, when $V=4$, DiT inputs decrease from 24 to 9, substantially lowering the conditional memory footprint compared to a symmetric baseline. 

To differentiate these varied inputs within the network, we leverage the 3D positional encoding native to the pre-trained image editing model, utilizing its first dimension to identify the condition type, thereby enabling the network to seamlessly integrate lighting references with global spatial awareness.

\subsubsection{Training Objective}

At each training iteration, we first sample a scene alongside a local subset of $V$ viewpoints. To formulate the training sample for relighting, we randomly select two distinct illuminations: a source illumination to provide the input images $\{I_i\}_{i=1}^V$, and a target illumination to render the ground-truth relit images $\{R_i\}_{i=1}^V$. Following our asymmetric conditioning scheme, we then uniformly sample an active view index $m \in \{1, \dots, V\}$. 

The model is optimized via a flow matching~\cite{lipman2022flow} objective. During the forward process, the ground-truth relit images $\{R_i\}_{i=1}^V$ are mapped to their latent representations $\{\mathbf{z}_0^i\}_{i=1}^V$ via a pre-trained VAE~\cite{KingmaVAE}. We then sample Gaussian noise $\mathbf{z}_1 \sim \mathcal{N}(\mathbf{0}, \mathbf{I})$ and construct the time-dependent noisy states via linear interpolation: $\mathbf{z}_t = (1-t)\mathbf{z}_0 + t\mathbf{z}_1$.
Conditioned on the active view's dense guiding signals, specifically the source image $I_m$, the spatial lighting condition $L_m$, and the illumination palettes $\{B_m\}$ separately encoded via $\mathcal{E}(\cdot)$, the network $v_\theta$ predicts the constant velocity field $u_t = \mathbf{z}_1^m - \mathbf{z}_0^m$ for the active view. The optimization objective is defined as:

\begin{equation}
    \mathcal{L} = \left\| v_\theta\left(\mathbf{z}_t^m, t, \underbrace{\mathcal{E}(I_m), \mathcal{E}(L_m), \{\mathcal{E}(B_m)\}}_{\text{active guiding signals}}, \underbrace{\{\mathbf{z}_t^k\}_{k \neq m}}_{\text{inactive views}}\right) - (\mathbf{z}_1^m - \mathbf{z}_0^m) \right\|^2
\end{equation}

Consistent with this design, gradients are back-propagated solely through the active target's prediction path, treating the inactive latents $\{\mathbf{z}_t^k\}_{k \neq m}$ purely as gradient-free spatial references.

\subsubsection{Multi-View Aware Inference}

During inference, we concurrently synthesize the full set of $N$ viewpoints. We initialize the process by drawing random Gaussian noise for the entire scene, denoted as $\mathcal{Z}_1 = \{\mathbf{z}_1^1, \dots, \mathbf{z}_1^N\}$.
At each denoising step $t$, we compute the velocity fields for all $N$ views. To achieve this, we formulate the prediction for each view $m \in \{1, \dots, N\}$ by setting it as the active view. For this active view $m$, we prepare its dense guiding signals, specifically the encoded source image $\mathcal{E}(I_m)$, spatial lighting condition $\mathcal{E}(L_m)$, and illumination palettes $\{\mathcal{E}(B_m)\}$, alongside a context window of $V-1$ inactive spatial anchors sampled from the remaining noisy state $\mathcal{Z}_t \setminus \{\mathbf{z}_t^m\}$. Conditioned on these combined inputs, the network $v_\theta$ predicts the velocity field $u_t^m$ exclusively for view $m$.
After computing this independent forward pass for all $N$ views, we aggregate the resulting velocity fields $\{u_t^1, \dots, u_t^N\}$ to form the joint velocity for the entire scene. The ODE solver then uses this joint velocity to update the global scene state $\mathcal{Z}_t$ a step backward toward $t=0$. Once the integration reaches $t=0$, the latents are decoded via the VAE to produce the final multi-view relit output $\mathcal{R}$.

\section{Experiments}
\label{sec:experiments}

\subsection{Implementation Details}
\label{sec:exp_setup}

Both the illumination distillation and illumination casting stages are built upon Flux.2 Klein 9B~\cite{flux-2-2025} to leverage its powerful generative priors.
The models in both stages are optimized using a flow matching objective~\cite{lipman2022flow} with AdamW~\cite{loshchilov2018decoupled}, and are trained for 1 epoch with a learning rate of $1 \times 10^{-4}$.
In the illumination distillation phase, we train LoRA modules with a rank of 32 and a batch size of 8.
In the illumination casting stage, we use a higher LoRA rank of 128 and a batch size of 32 to handle the more complex spatial conditions.
We train the casting stage on paired synthetic multi-view relighting data rendered under diverse lighting configurations, using 480 scenes for training and 10 held-out scenes for evaluation.
During casting-stage training, we randomly sample a local subset of $V=4$ viewpoints to form the context window.

\subsection{Comparison with Baseline Methods}
\begin{figure}[t]
\centering
\includegraphics[width=\linewidth]{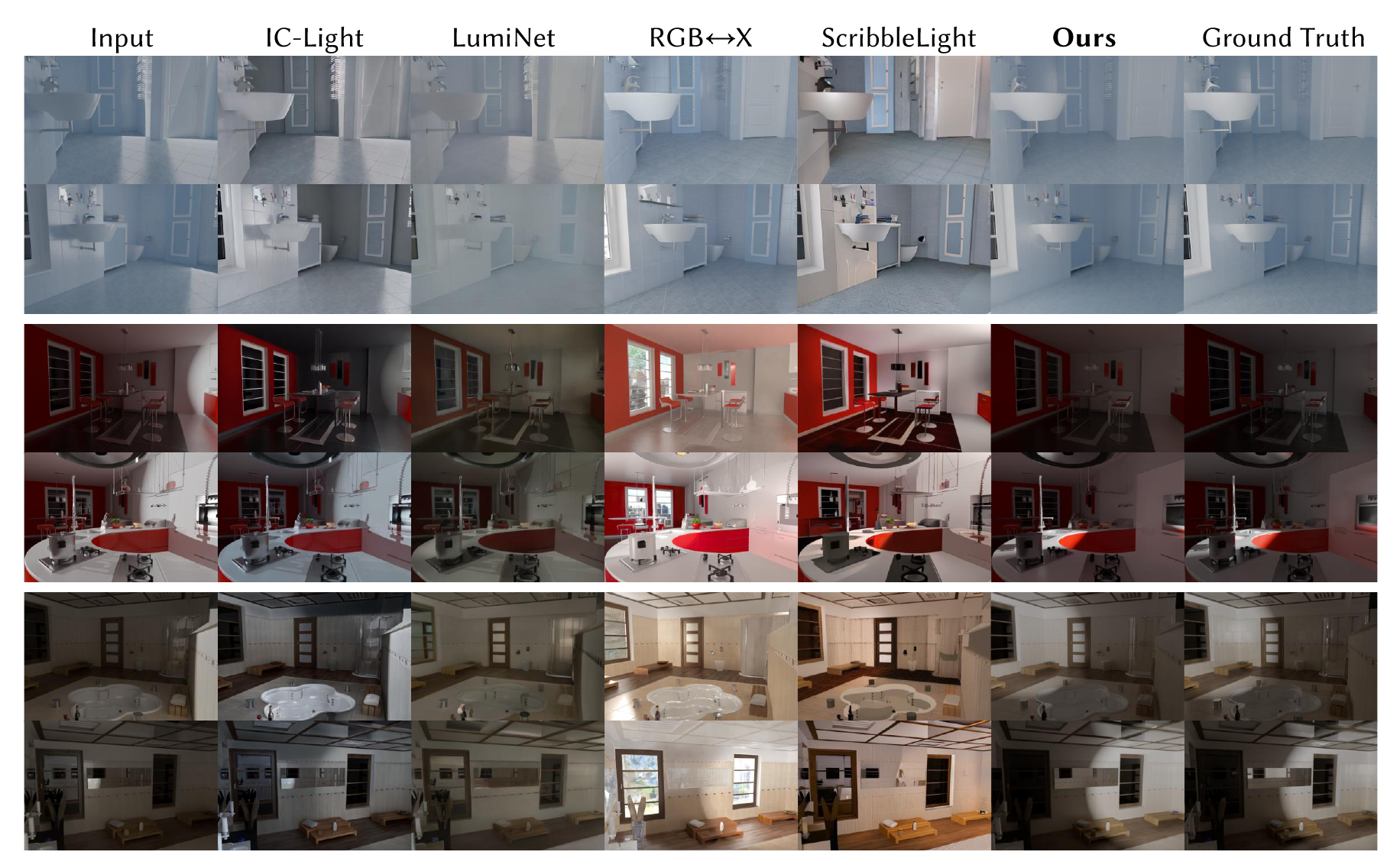}
\caption{Qualitative comparison against baselines on synthetic scenes. Our method achieves precise spatial control, realistic appearance, and multi-view consistency.}
\label{fig:qualitative_sota}
\end{figure}

\begin{table}[t]
\centering
\caption{Quantitative comparison against baselines on synthetic scenes, including image fidelity and geometry-aware multi-view consistency.}
\label{tab:relight_metrics}
\begin{tabular}{lccccc}
\toprule
 & RGB$\leftrightarrow$X~\cite{zeng2024rgb} & IC-Light~\cite{zhang2024iclight} & LumiNet~\cite{xing2025luminet} & ScribbleLight~\cite{choi2025scribblelight} & Ours \\
\midrule
PSNR $\uparrow$    & 10.497 & 13.982 & 16.593 & 12.591 & \textbf{24.453} \\
SSIM $\uparrow$    & 0.525  & 0.622  & 0.607  & 0.504  & \textbf{0.872}  \\
LPIPS $\downarrow$ & 0.394  & 0.281  & 0.399  & 0.431  & \textbf{0.126}  \\
MV MSE $\downarrow$ & 0.0227 & 0.0184 & 0.0159 & 0.0218 & \textbf{0.0090} \\
MV LPIPS $\downarrow$ & 0.176 & 0.143 & 0.195 & 0.208 & \textbf{0.115} \\
\bottomrule
\end{tabular}
\end{table}

We evaluate Lume-Palette against recent open-source relighting methods, including RGB$\leftrightarrow$X~\cite{zeng2024rgb}, IC-Light~\cite{zhang2024iclight}, LumiNet~\cite{xing2025luminet}, and ScribbleLight~\cite{choi2025scribblelight}. We first evaluate relighting accuracy on diverse synthetic scenes. We report image-fidelity metrics, including PSNR, SSIM, and LPIPS. In addition, we evaluate geometry-aware multi-view consistency by warping each generated relit view to its adjacent view using depth and camera poses, and computing MSE and LPIPS against the corresponding generated adjacent-view result in the overlapping region. As shown in Table~\ref{tab:relight_metrics}, Lume-Palette achieves the best results among the evaluated baselines under the adopted metrics. Qualitative comparisons in Fig.~\ref{fig:qualitative_sota} further illustrate the behavior of different methods under diverse spatial lighting configurations. IC-Light often struggles to disentangle and remove the baked-in source illumination, while LumiNet shows limited ability to impose complex target illumination. Methods relying on explicit intrinsic decomposition, such as RGB$\leftrightarrow$X and ScribbleLight, are sensitive to albedo prediction errors and may introduce noticeable color shifts. In comparison, Lume-Palette produces spatially aligned relighting results with visually plausible appearance and improved cross-view coherence. It captures a range of lighting effects, from high-frequency spotlight shadows to softer illumination from rectangular area lights, while keeping the resulting shading patterns coherent across viewpoints.

\begin{figure}[t]
\centering
\includegraphics[width=\linewidth]{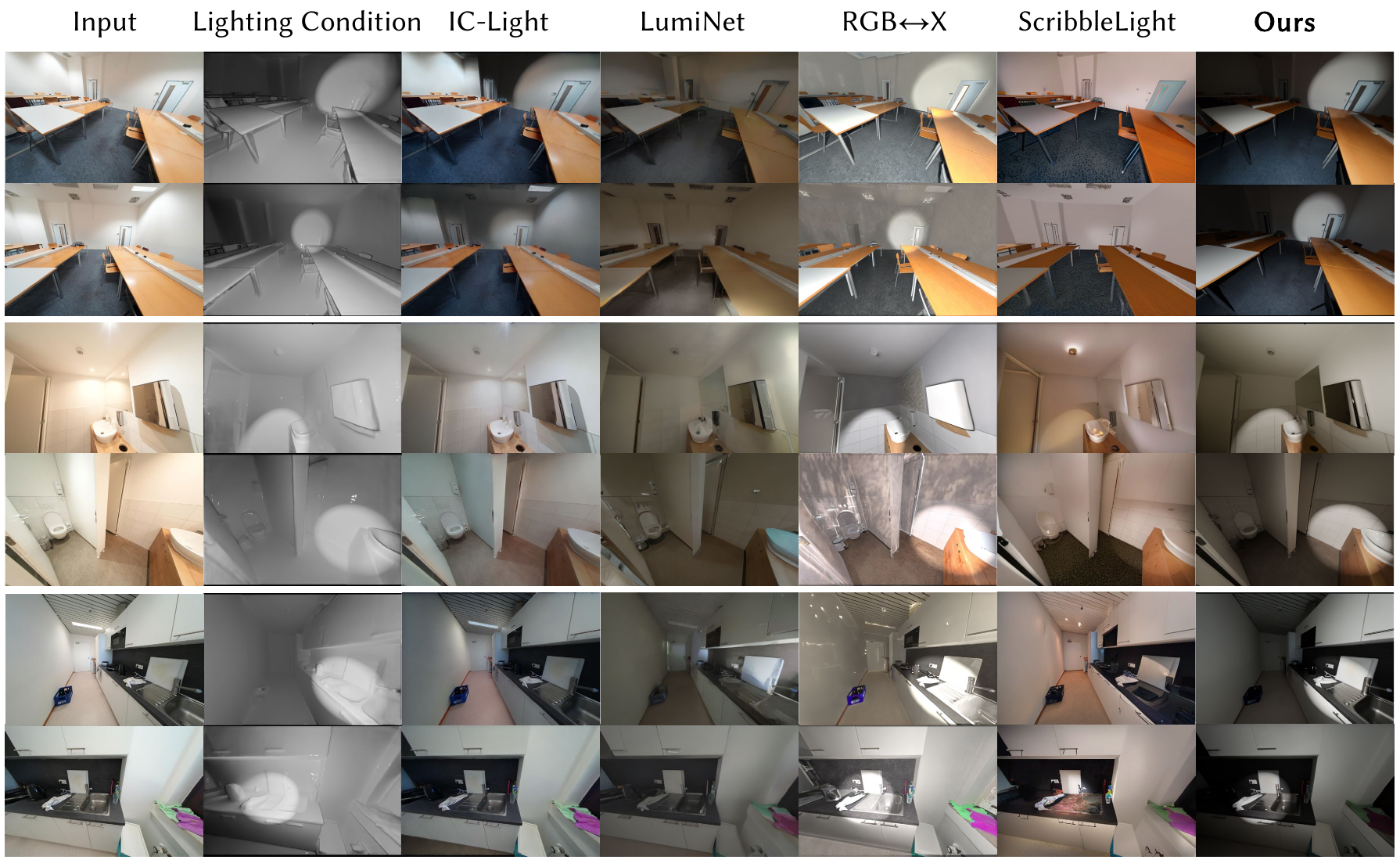}
\caption{Qualitative comparison against baselines on real scenes. Lume-Palette robustly synthesizes realistic shading that adheres to imposed spatial lighting conditions.}
\label{fig:qualitative_sota_real}
\end{figure}

\begin{table}[t]
\centering
\caption{Real-scene user study results using Bradley--Terry scores. Higher scores indicate stronger user preference.}
\label{tab:user_study_bt}
\begin{tabular}{lccc}
\toprule
Method & Realism $\uparrow$ & Lighting adherence $\uparrow$ & MV consistency $\uparrow$ \\
\midrule
RGB$\leftrightarrow$X & -1.498 & 0.807 & -0.038 \\
IC-Light & 1.760 & -1.031 & -0.038 \\
LumiNet & -0.571 & -1.981 & -0.542 \\
ScribbleLight & -1.742 & -0.596 & -0.895 \\
Ours & \textbf{2.051} & \textbf{2.801} & \textbf{1.514} \\
\bottomrule
\end{tabular}
\end{table}

We further evaluate Lume-Palette on real scenes from ScanNet++~\cite{yeshwanth2023scannet++}. As shown in Fig.~\ref{fig:qualitative_sota_real}, Lume-Palette can synthesize plausible shading that follows the imposed spatial lighting conditions in complex real scenes. The results show spatially varying illumination effects rather than global tone changes, with brighter and darker regions appearing in accordance with the provided lighting maps. Since ground-truth relighting is unavailable for real scenes, we conduct a user study with 25 participants to assess perceptual quality. In each trial, participants compare two anonymized method outputs according to realism, lighting adherence to the reference diffuse lighting condition, and multi-view consistency. We collected 150 valid pairwise preference judgments spanning different scenes, method pairs, and evaluation criteria. The preferences are aggregated using a Bradley--Terry model, and the resulting scores are reported in Table~\ref{tab:user_study_bt}. Our method is consistently preferred across all three dimensions.

\subsection{Ablation Studies}
\label{sec:ablation}
\begin{figure}[t]
\centering
\includegraphics[width=\linewidth]{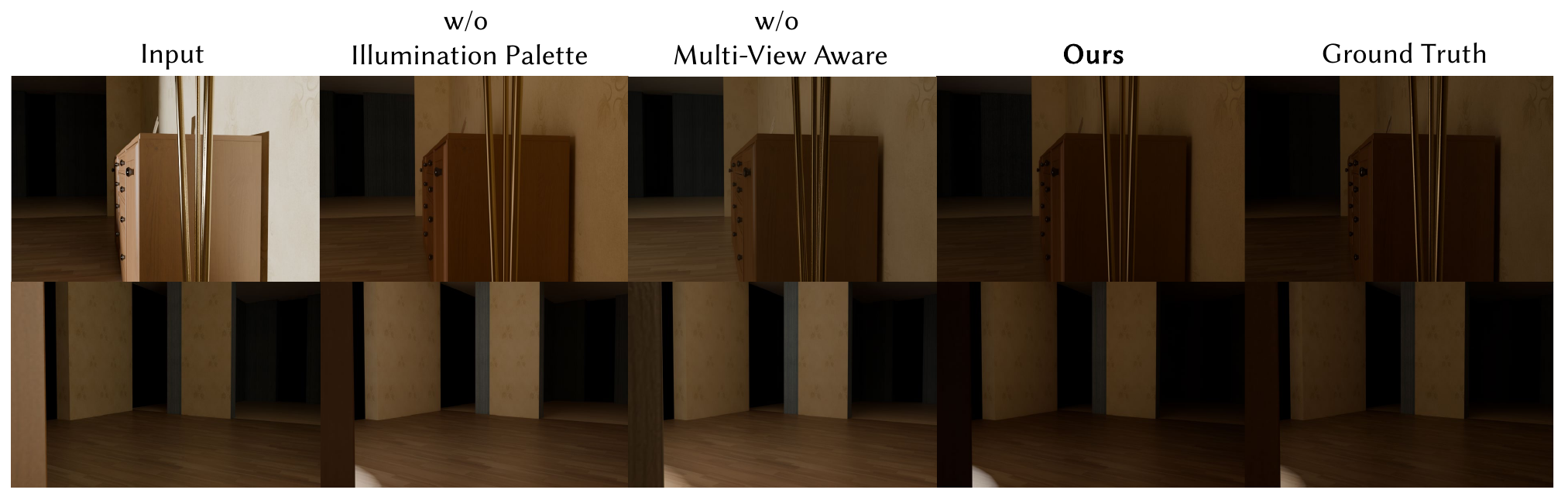}
\caption{Qualitative ablation of the illumination palette and multi-view awareness. Removing the illumination palette or multi-view awareness degrades performance.}
\label{fig:ablation_ref}
\end{figure}

\begin{figure}[t]
\centering
\includegraphics[width=\linewidth]{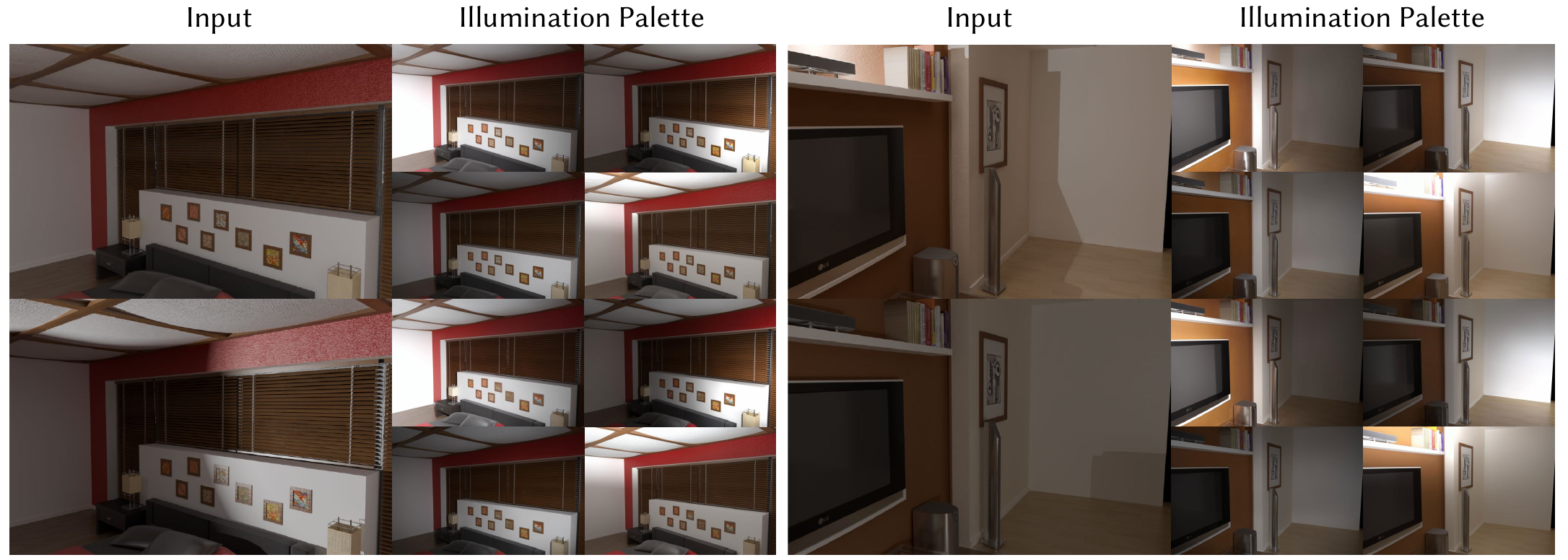}
\caption{Robustness of the illumination palette to varying input illumination. Changes in input light sources and shadows have almost no effect on the extracted palette.}
\label{fig:camp_side}
\end{figure}

\begin{table}[t]
\centering
\caption{Quantitative ablation of illumination palettes and multi-view awareness.}
\label{tab:ablation-mv-ip}
\begin{tabular}{lccc}
\toprule
Model Variant & PSNR $\uparrow$ & SSIM $\uparrow$ & LPIPS $\downarrow$ \\
\midrule
w/o Multi-View Aware \& Illumination Palette & 21.973 & 0.834 & 0.140 \\
w/o Multi-View Aware & 23.618 & 0.864 & 0.131 \\
w/o Illumination Palette                       & 22.748 & 0.854 & 0.128 \\
\midrule
\textbf{Full Lume-Palette}                     & \textbf{24.453} & \textbf{0.872} & \textbf{0.126} \\
\bottomrule
\end{tabular}
\end{table}
\begin{figure}[t]
\centering
\includegraphics[width=\linewidth]{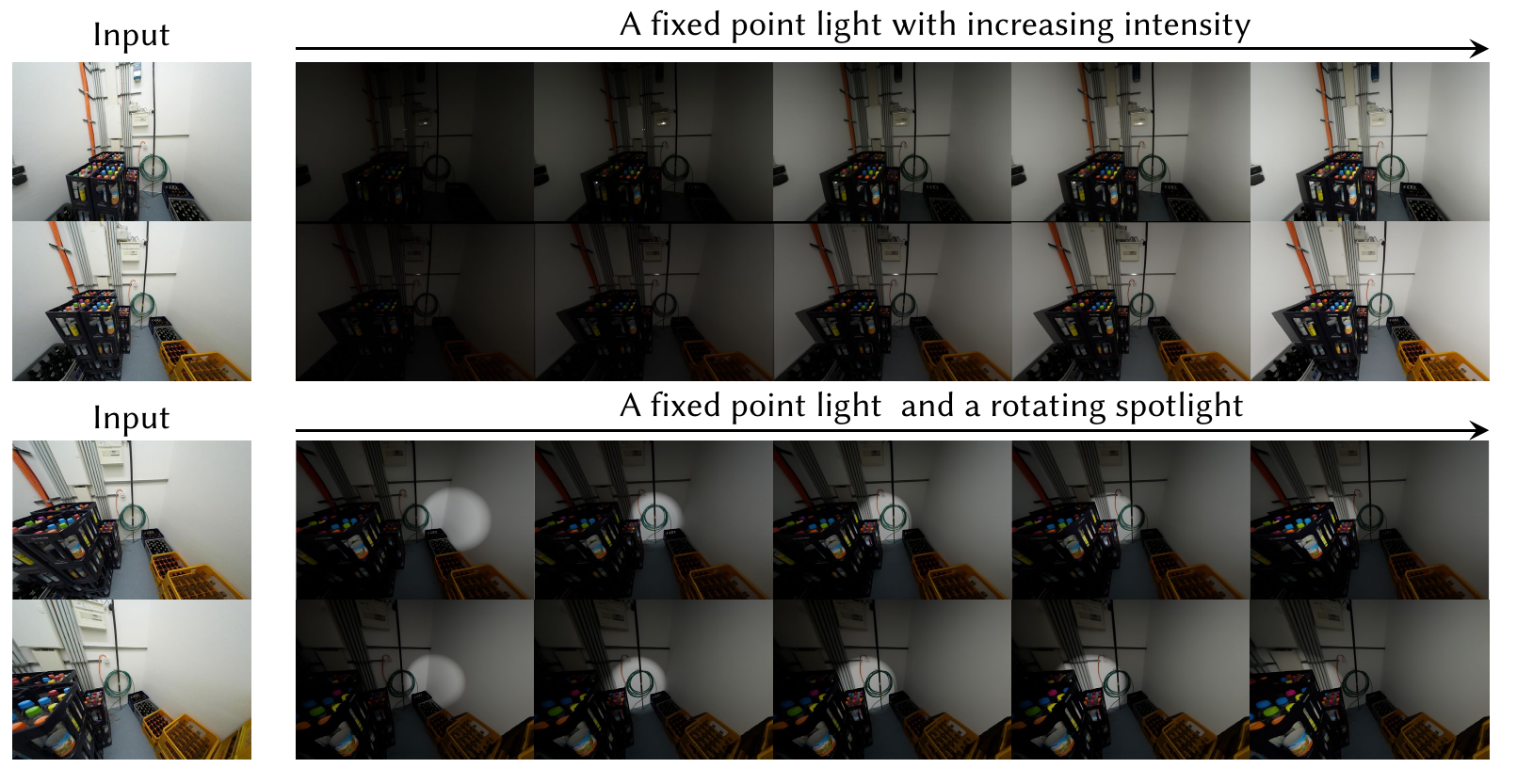}
\caption{Continuous spatial and dynamic controllability of light intensity and direction. The generated illumination, including specular highlights and cast shadows, updates naturally and remains firmly attached to the underlying geometry.}
\label{fig:continious}
\end{figure}
We conduct ablation studies to validate the core components of our proposed framework.

\textbf{Effectiveness of Illumination Distillation.} To validate the effectiveness of illumination distillation, we train a variant (w/o Illumination Palette) that directly conditions the image editing model on explicit spatial maps without the intermediate illumination palettes. As shown in Table~\ref{tab:ablation-mv-ip}, this direct enforcement degrades the model's generative prior, leading to a noticeable drop in PSNR and a worse LPIPS score compared to the full model. Qualitatively, Fig.~\ref{fig:ablation_ref} shows that removing this component leads to noticeable synthetic artifacts and a loss of intrinsic material properties. Furthermore, Fig.~\ref{fig:camp_side} demonstrates the robustness of the Illumination Palette, showing that it can consistently extract canonical illumination regardless of varying initial input lighting. This stable representation provides reliable appearance priors for the subsequent casting stage.

\textbf{Impact of Asymmetric Multi-View Conditioning.} By removing the inactive spatial anchors, we train a single-view variant (w/o Multi-View Aware). As demonstrated quantitatively and qualitatively, the overall relighting quality of this baseline is inferior to our full Lume-Palette. Fundamentally, multi-view inputs provide cross-view spatial anchors and parallax cues, helping the model maintain consistent light-surface interactions across viewpoints. Deprived of this multi-view spatial awareness, the model generates noticeably flatter shading and less accurate light-surface interactions. This confirms that our asymmetric conditioning scheme is essential for deeply understanding scene materials, thereby enhancing the overall photorealism of the final relighting results more faithfully.

\subsection{Spatial and Dynamic Controllability}
\begin{figure}[t]
\centering
\includegraphics[width=\linewidth]{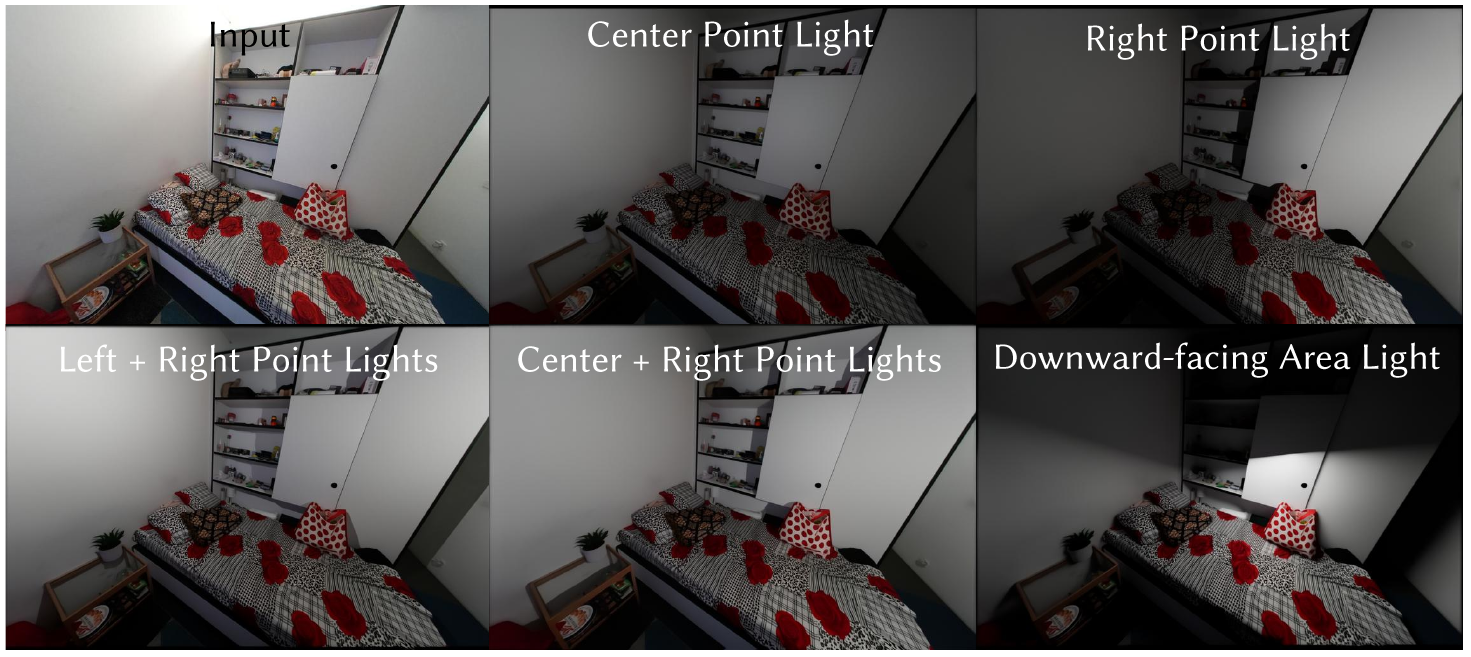}
\caption{Lume-Palette supports flexible lighting configurations, including point lights at different positions, multiple simultaneous lights, and rectangular area lights.}
\label{fig:diverse_lighting}
\end{figure}
A core advantage of Lume-Palette is its fine-grained lighting control. As illustrated in Fig.~\ref{fig:continious}, our framework supports continuous spatial and dynamic controllability beyond discrete lighting presets. Users can adjust the intensity of a fixed point light or rotate a spotlight within the 3D space, and the relit appearance changes progressively with the edited light. The generated illumination, including specular highlights, cast shadows, and local brightness falloff, updates naturally and remains attached to the underlying geometry.

Beyond single-light editing, Lume-Palette can handle more complex lighting layouts. Fig.~\ref{fig:diverse_lighting} shows results under diverse setups, including point lights at different positions, multiple lights, and a rectangular area light. These examples demonstrate that the receiver-centric lighting representation can express different source types and spatial arrangements within the same framework. Multiple lights jointly affect the scene with spatially varying contributions, while area lighting produces broader and softer shading patterns than point lighting.

\section{Conclusion and Limitations}
We presented Lume-Palette, a progressive framework for spatially controllable multi-view indoor scene relighting. By decoupling relighting into illumination distillation and illumination casting, our method leverages canonical illumination palettes to preserve realistic material--light interactions, while using receiver-centric spatial lighting conditions to support explicit 3D light placement. To enable efficient joint multi-view generation, we further introduced an asymmetric conditioning strategy that provides dense guidance only to the active view while maintaining cross-view spatial anchors through inactive latents. Experiments on synthetic and real-world scenes show that Lume-Palette produces photorealistic, spatially aligned, and multi-view consistent relighting results, comparing favorably with baseline methods under the adopted metrics.

Despite these encouraging results, our method still has several limitations. Since receiver-centric lighting conditions are rendered from coarse reconstructed geometry, inaccurate geometry may degrade relighting quality, especially around thin structures, reflective or transparent materials, and scenes with complex indirect illumination. We hope this limitation motivates future work on more robust geometry-aware conditioning for controllable multi-view relighting.

\section*{Acknowledgements}
This study was supported in part by the Centre for Perceptual and Interactive Intelligence (CPII) Ltd., a CUHK-led InnoCentre under the InnoHK initiative of the Innovation and Technology Commission of the Hong Kong Special Administrative Region Government. This study was supported by CUHK-CUHK(SZ)-GDSTC Joint Collaboration Fund No. 2025A0505000053. We would like to thank the reviewers for their insightful comments.

\bibliographystyle{splncs04}
\bibliography{main}

\end{document}